%% file: main.tex
\documentclass[10pt,twocolumn,letterpaper]{article}
\pdfoutput=1 

\usepackage{cvpr}
\usepackage{times}
\usepackage{epsfig}
\usepackage{graphicx}
\usepackage{amsmath}
\usepackage{amssymb}

\usepackage{enumitem}
\usepackage{subfigure} 
\setlist[itemize]{leftmargin=*}
\usepackage{amssymb}
\usepackage{pifont}
\usepackage{makecell}
\setcellgapes{4pt}
\usepackage{multirow}
\usepackage[breaklinks=true,bookmarks=false,colorlinks,linkcolor=red]{hyperref}
\usepackage[dvipsnames]{xcolor}
\usepackage{bbm}

\usepackage[breaklinks=true,bookmarks=false,colorlinks,linkcolor=red]{hyperref}

\cvprfinalcopy 

\def\cvprPaperID{****} 

\newcommand{\myparagraph}[1]{\vspace{3pt}\noindent{\bf #1 \ }}

\newcommand*{\mybox}[1]{\framebox{#1}}

\begin{document}

\title{Boundary-Aware Dense Feature Indicator for Single-Stage 3D Object Detection from Point Clouds}

\author{
Guodong Xu$^{1,2}$, Wenxiao Wang$^{1,2}$, Zili Liu$^{1,2}$, Liang Xie$^{1,2}$, Zheng Yang$^2$, Haifeng Liu$^1$, Deng Cai$^{1,2}$\\\
$^1$State Key Lab of CAD\&CG, Zhejiang University, Hangzhou, China\\
$^2$Fabu Inc., Hangzhou, China\\
}

\maketitle

\input{sec_abstract}
\input{sec_introduction}
\input{sec_related_work}
\input{sec_main}
\input{sec_experiments}
\input{sec_conclusion}

{\small
\bibliographystyle{ieee_fullname}
\bibliography{egbib}
}

\end{document}

%% file: sec_abstract.tex
\begin{abstract}
3D object detection based on point clouds has become more and more popular. Some methods propose localizing 3D objects directly from raw point clouds to avoid information loss. However, these methods come with complex structures and significant computational overhead, limiting its broader application in real-time scenarios. Some methods choose to transform the point cloud data into compact tensors first and leverage off-the-shelf 2D detectors to propose 3D objects, which is much faster and achieves state-of-the-art results. However, because of the inconsistency between 2D and 3D data, we argue that the performance of compact tensor-based 3D detectors is restricted if we use 2D detectors without corresponding modification. Specifically, the distribution of point clouds is uneven, with most points gather on the boundary of objects, while detectors for 2D data always extract features evenly. Motivated by this observation, we propose DENse Feature Indicator (DENFI), a universal module that helps 3D detectors focus on the densest region of the point clouds in a boundary-aware manner. Moreover, DENFI is lightweight and guarantees real-time speed when applied to 3D object detectors. Experiments on KITTI dataset show that DENFI improves the performance of the baseline single-stage detector remarkably, which achieves new state-of-the-art performance among previous 3D detectors, including both two-stage and multi-sensor fusion methods, in terms of mAP with a 34FPS detection speed.






\end{abstract}


%% file: sec_introduction.tex
\vspace*{-0.4cm}
\section{Introduction}
\label{sec:introduction}

\begin{figure}[t]
    \includegraphics[width=0.48\textwidth]{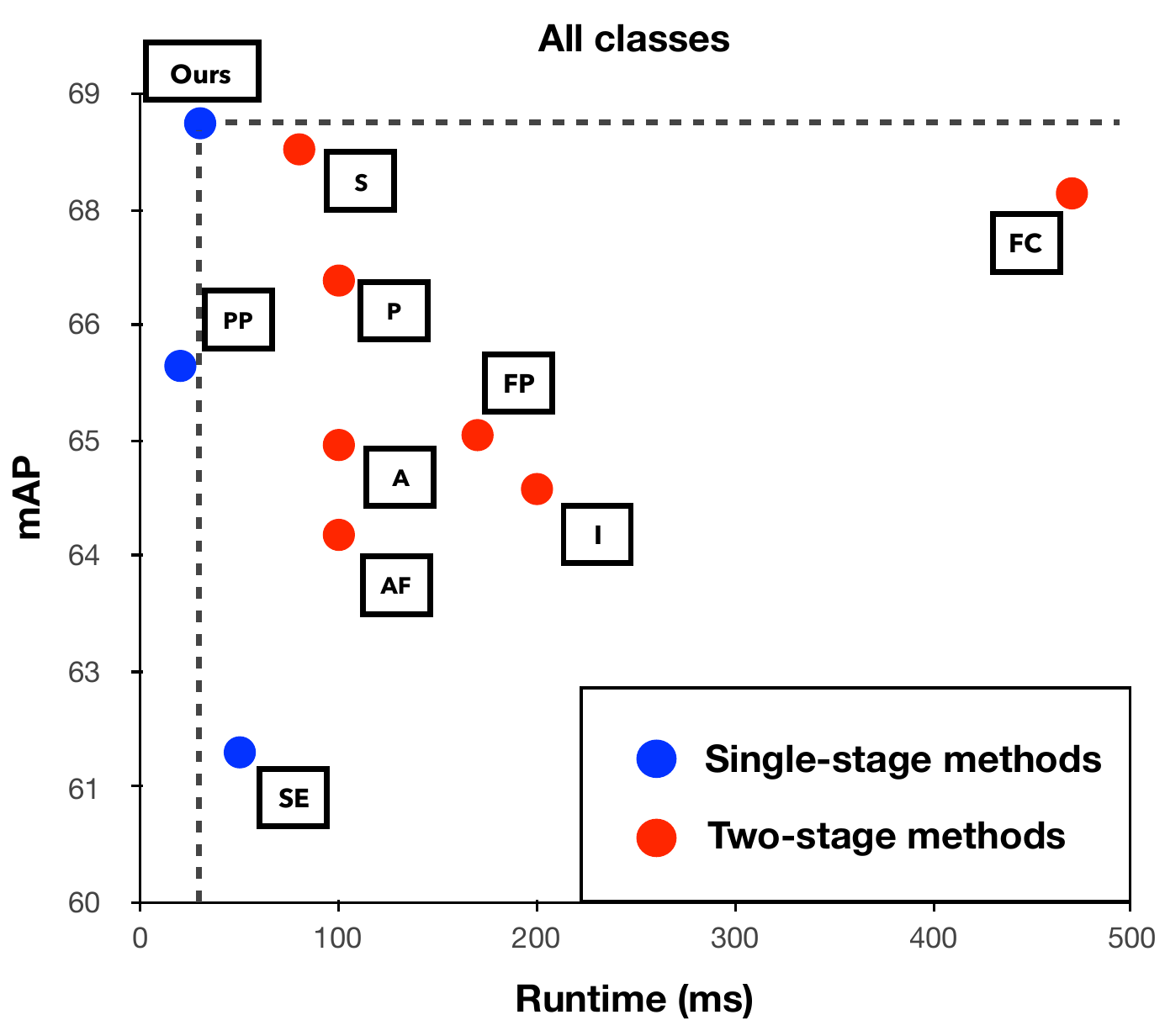}
    \centering    
    \caption{Bird’s eye view performance on the officially most critical \cite{geiger2012we} “moderate” difficulty level versus Speed (ms) on KITTI \emph{test} split for our method: \mybox{Ours}. We also show state-of-the-art methods from the KITTI leaderboard: \mybox{PP}: PointPillars\cite{lang2019pointpillars}, \mybox{S}: STD \cite{yang2019std}, \mybox{P}: PointRCNN \cite{shi2019pointrcnn}, \mybox{A}: AVOD \cite{ku2018joint}, \mybox{AF}: AVOD-FPN \cite{ku2018joint}, \mybox{I}: IPOD \cite{yang2018ipod}, \mybox{FP}: F-PointNet \cite{qi2018frustum}, \mybox{FC}: F-ConvNet \cite{wang2019frustum} \mybox{SE}: SECOND \cite{yan2018second}. Our method outperforms all 3D detectors, including both two-stage and multi-sensors based, by a large margin in terms of mAP while running at 34FPS. Detailed results are shown in Table \ref{table:experiments-kitti-test}.}
    \label{fig:method}
\end{figure}

\begin{figure*}
    \centering
    \includegraphics[width=1.0\textwidth]{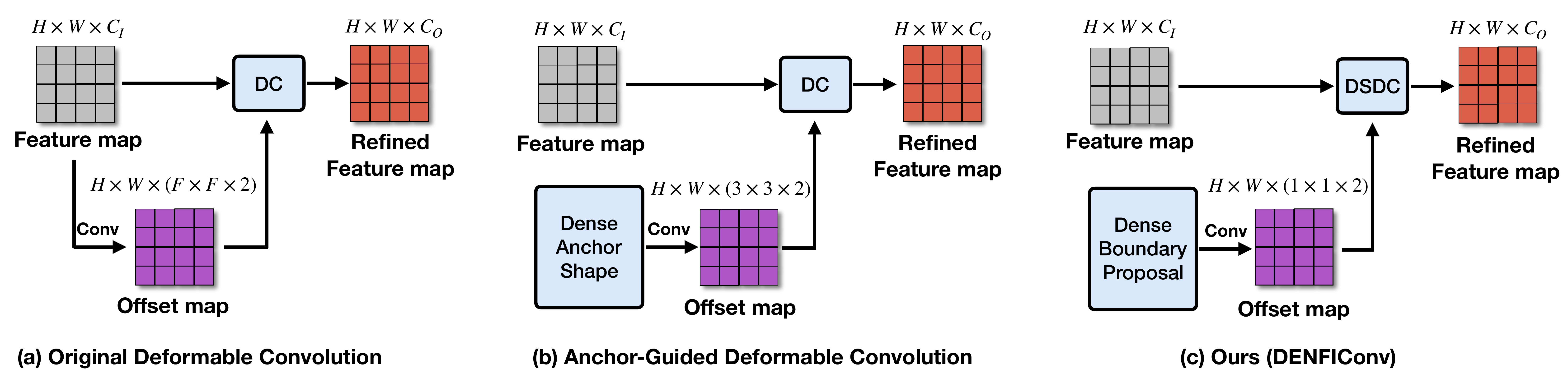}
    \centering    
    \caption{{Different Deformable-Convolution-Based Feature Extractor.} DC denotes deformable convolution, and DSDC denotes depth-wise separable deformable convolution (See Section \ref{subsec:fcaf-fc}). (a) utilizes the feature map itself to learn the offsets of the convolution kernel at each position on the feature map, where $F$ denotes the size of the convolution kernel, and $2$ denotes the offset of convolution sampling point along $H, W$ direction respectively. (b) attempts to align features for learned anchors by learning offsets from dense anchor shape. (c) is the proposed DENFIConv, which capture dense features in an boundary-aware manner by learning offsets from the dense boundary proposal.
}
    \label{fig:related-dconv}
\end{figure*}


With the rapid development of depth sensors and application requirements in the fields of auto-driving and robotics, point clouds based 3D object detection has become more and more popular. Many detection methods have been proposed, and these methods could be divided into two categories: point-based and tensor-based.

Point-based methods \cite{shi2019pointrcnn,yang2019std, chen2019fast} do 3D detection by using raw point clouds directly to avoid loss of geometric details. However, these architectures come with complex structures and significant computational overhead, making it unfavorable for real-time scenarios with large-scale point clouds.

Unlike point-based methods, tensor-based methods transform point clouds into tensors such as 3D voxel grids or bird's-eye-view (BEV) images \cite{zhou2018voxelnet,yang2018pixor,yan2018second,lang2019pointpillars,yang2018hdnet} first, then localize 3D objects with mature 2D detectors like SSD \cite{liu2016ssd}, RetinaNet \cite{lin2017focal} and so on. Tensor-based methods take advantage of compact and fast architectures in 2D detection, which achieves great success. 

However, there is still one problem in tensor-based methods: the performance of 2D detectors is restricted because of the inconsistency between 2D and 3D data. Specifically, the distribution of point clouds is uneven, and the densest regions, together with its corresponding feature (what we call dense features),  that are important for localizing objects gather on only a small part of the 3D space. However, the 2D detector always extracts features evenly, so it has difficulty aggregating dense features for accurate detection.

In order to capture dense features, one possible way is to increase the receptive field, which, however, may introduce more inclusion of nearby objects and clutter. An alternative way would be to adopt deformable convolution \cite{dai2017deformable}, which tries to capture the most critical features by inducing ``offset'' into the convolution layer. However, ``offset'' in deformable convolution is also learned from feature maps, which is sub-optimal because it ignores the inherent distribution of point clouds and thus can only be optimized implicitly.

Note that most of the point clouds gather on the boundary of objects. To better utilize the dense features lying around the boundary, we propose our module called DENse Feature Indicator (DENFI) to capture dense features explicitly in a boundary-aware manner. The steps are as follows: \textbf{\texttt{1}}) DENFI utilizes a Dense Boundary Proposal Module (DBPM) to predicts the dense boundary information according to feature maps extracted by the backbone. \textbf{\texttt{2}}) DENFI leverages DENFIConv to adaptively captures dense features from the backbone feature map under the guidance of the dense boundary information from DBPM. The refined features are then used for detection purposes. 



Extensive experiments on the challenging benchmark of KITTI \cite{geiger2012we} dataset have shown that DENFI improves the performance of single-stage 3D detector PointPillars \cite{lang2019pointpillars} by \textbf{2.71, 2.70} and \textbf{2.13 mAP} on objects with easy, moderate and hard difficulty level. Also, DENFI is lightweight and permits the detectors to run at real-time speed.



It is worthwhile to highlight our contributions:

\begin{itemize}
    \item We point out the inconsistency of adopting 2D detection frameworks for 3D object detection due to uneven distribution of point clouds, and it can be mitigated if we adaptively capture dense features from uneven distribution in a boundary-aware manner.
    \item We propose an efficient and universal module called DENFI, which is specially designed for 3D detection to help 3D detectors capture dense features. 
    \item We propose DENFIDet by combining DENFI with PointPillars, which achieves new state-of-the-art performance on KITTI dataset while running at 34FPS. 
\end{itemize}



%% file: sec_related_work.tex
\vskip 0px
\section{Related Work}
\label{sec:related-work}

\myparagraph{Point-Based 3D Object Detection from Point Clouds} 
Current Point-Based detectors heavily rely on a two-stage framework to achieve decent performance. \cite{shi2019pointrcnn} propose PointRCNN, a two-stage network, for directly detecting 3D objects from the raw point clouds. It first performs points segmentation to localize foreground points and regresses proposals from foreground points. Then it refines the proposals to generate final detection results by further gathering local spatial features. STD \cite{yang2019std} introduces spherical anchors at the first stage of PointRCNN to reduce the number of foreground points. It then applies voxelization at the second stage of PointRCNN for speedup, which integrates the speed advantage of the tensor-based methods into point-based methods. However, since this type of work direct processes the raw point clouds, it is difficult to be extended to large-scale point clouds scenarios requiring real-time speed. 



%


\begin{figure*}[t]
    \includegraphics[width=1\textwidth]{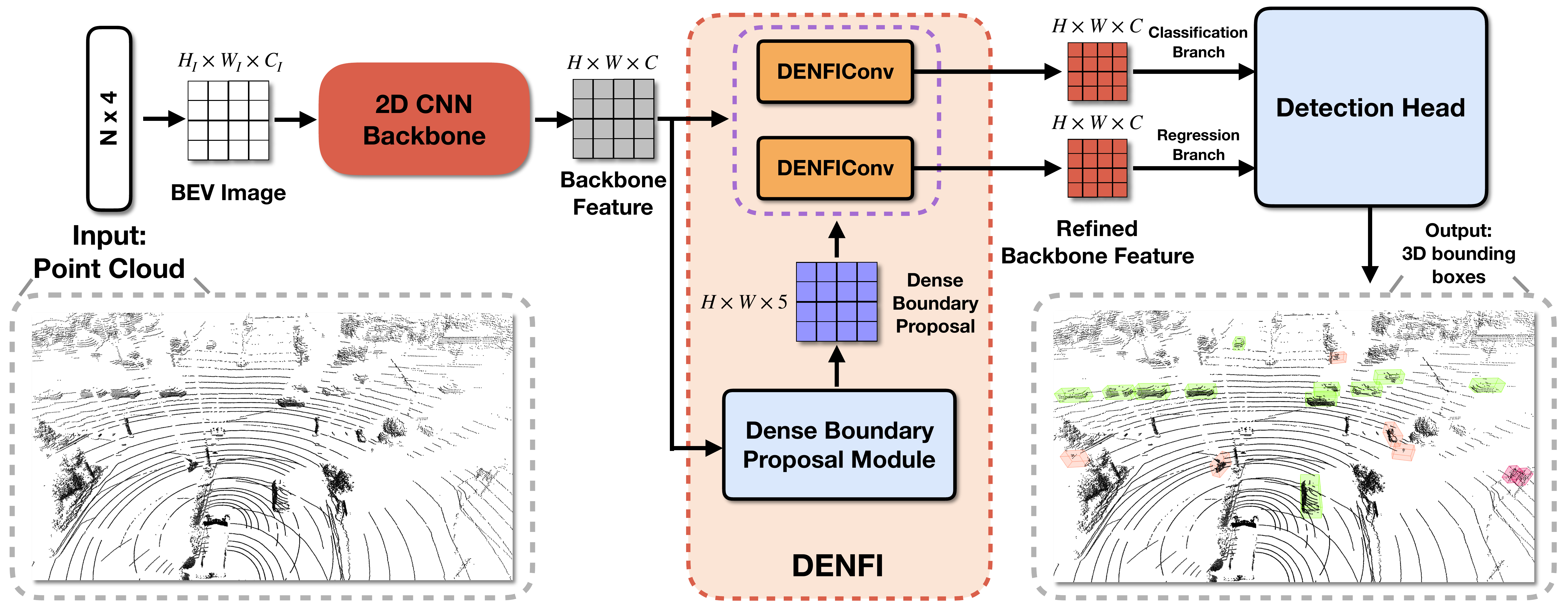}
    \centering    
    \caption{{Illustration of the DENFIDet architecture.} We insert DENFI (dashed orange box) into a common single-stage 3D detection framework. Given the feature extracted from the CNN backbone, DENFI generates dense boundary proposals with DBPM. These dense boundary proposals are used to guide DENFIConv capture dense features. Then the detection head leverages the refined feature to generate the final proposals. Note that BEV (Bird’s-Eye-View) Image is a 2D compact tensor representation of the point cloud.}
    \label{fig:method}
\end{figure*}

\myparagraph{Tensor-Based 3D Object Detection from Point Cloud} 
Tensor-based detection transforms the point cloud into a compact tensor first and leverages off-the-shelf 2D detectors to this tensor. So, tensor-based detection, just like 2D object detection, can be divided into two categories: \textbf{anchor-based detection} and \textbf{anchor-free detection}: \textbf{\texttt{1}}) \textbf{anchor-based detection} relies on a set of predefined anchor boxes. \cite{zhou2018voxelnet} proposes VoxelNet, an end-to-end network unifying feature extraction and bounding box prediction into a single stage. \cite{yan2018second} explores the use of sparse convolution \cite{graham2017submanifold,graham20183d} to efficiently accelerate VoxelNet. \cite{lang2019pointpillars} further proposed PointPillars to learn the feature suitable for 2D convolution, making the network fast and accurate.  \textbf{\texttt{2}}) \textbf{anchor-free detection} predicts the position of bounding boxes directly, and there is no need to compute the intersection over-union (IOU) scores between all anchor boxes and ground-truth boxes during training, thus making it simple and time-efficient, which is favorable for industrial application. \cite{yang2018pixor} proposed PIXOR as an end-to-end anchor-free network for speed and simplicity. \cite{yang2018hdnet} further improved PIXOR results by exploring high-precision maps information. Though simple and fast, anchor-free detection often performs worse than an anchor-based one on accuracy.

\myparagraph{Deformable-Convolution-Based Feature Extractor} 
\cite{dai2017deformable} proposes deformable convolution, which introduces offsets to the regular grid sampling locations in the standard convolution to enable more flexible feature extraction, and offsets are induced as learnable parameters learned from the feature map. Instead of learning offsets implicitly from the feature map, Guided-Anchoring \cite{wang2019region} propose learning offsets from the shape of learned anchors to align learned anchors with the feature map. Different from those methods, our DENFI learns offsets from the predicted boundary because of the characteristics of the point clouds as we described in the introduction. Besides, we design an operator called depthwise separable deformable convolution as an alternative to vanilla deformable convolution, which runs several times faster and delivers similar performance. We compare different Deformable-Convolution-Based Feature Extractor in Figure \ref{fig:related-dconv}.

%% file: sec_main.tex
\section{DENse Feature Indicator}
\label{sec:method}

To solve the problem we depicted in the introduction, we propose our DENse Feature Indicator (DENFI). DENFI is a fully convolutional module which can be used to extract dense features in a boundary-aware manner. As shown in Figure \ref{fig:method}, DENFI consists of two parts: dense boundary proposal module (DBPM), which predicts the boundary of objects, and DENFIConv, which extracts dense features.

\begin{figure}
    \centering
    \begin{minipage}[t]{1.0\linewidth}
        \includegraphics[width=1\linewidth]{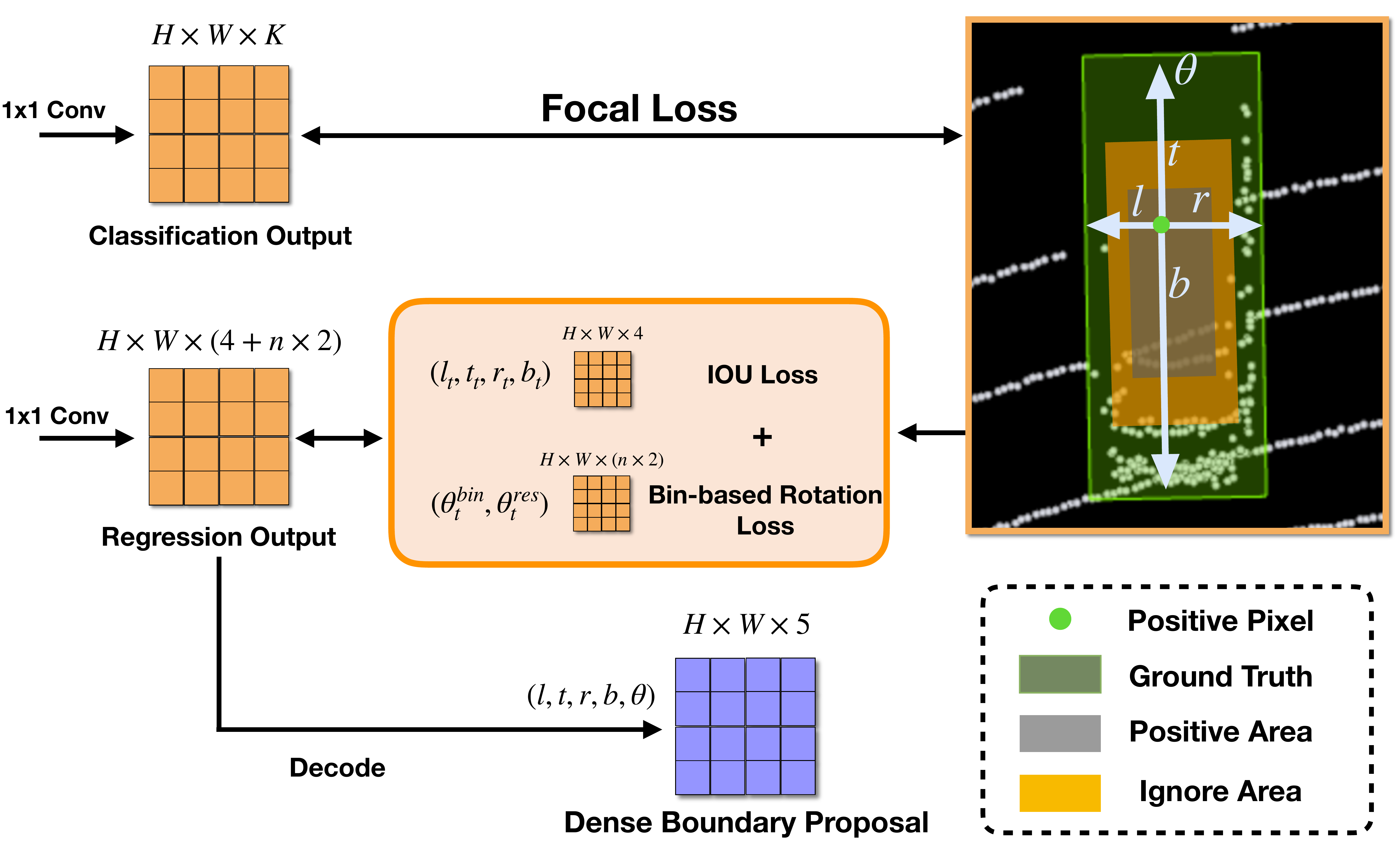}
    \end{minipage}%
    \caption{{Illustration of DBPM Design.} We extract the dense boundary proposal from the regression output for boundary indication, which contains $l, t, r, b, \theta$. The pixel in the positive area is defined as positive pixel.}
    \label{fig:method-anchor-free}
\end{figure}

\subsection{Dense Boundary Proposal Module}
\label{subsec:fcaf-af}
Since the point cloud is mostly distributed at the boundary of objects, we resort to the boundary of objects for dense features indication. We devise a simple but effective dense boundary proposal module (DBPM) for boundary proposal predictions, as shown in Figure \ref{fig:method-anchor-free}. It takes the backbone feature $F$ of size $H \times W \times C$ as its input and works in an anchor-free manner.

DBPM consists of two branches, which are the classification branch and the regression branch. The classification branch predicts the class score for each pixel of the feature map, and the regression branch performs boundary regression for each pixel of the feature map.

 

\myparagraph{Classification Branch}
The classification branch consists of a single 1x1 convolution, and its output channel number is equal to the number of categories $K$. As a result, the classification branch produces the probability map $\mathcal{P}$ of size $H \times W \times K$. The pixel $p_{i,j,k}$ represents the predicted score (with a range from 0 to 1) for the $k$th category, and a more significant value indicates a higher confidence. We also define the ground truth class for the pixel location $(i, j)$ as $k_{i, j}$.

For the definition of positive and negative samples, previous work \cite{huang2015densebox} proposed the "effective zone" for balanced sampling, which, however, only works on axis-aligned 2D bounding boxes. We extend it to rotated bounding boxes for 3D object detection in an anchor-free manner, as shown in Figure \ref{fig:method-anchor-free} (A \textit{Car} example). First, we define a rotated ground-truth box from the bird's-eye-view as $(x, y, w, l, \theta)$, where $(x,y)$ represents the center of the object, $(w, l)$ represents the size of the object, and $\theta$ represents the heading angle, which is within the range $[-\pi, \pi]$. The positive area $A_p$ for it is defined as a shrink version of the rotated ground-truth box $(x, y, \sigma_1 w, \sigma_1 l, \theta)$, and $\sigma_1$ is the positive scaling factor. Next, for the negative area $A_n$, we define another shrink version of the rotated ground-truth box $(x, y, \sigma_2 w, \sigma_2 l, \theta)$, where $\sigma_2$ is the negative scaling factor and $\sigma_1 < \sigma_2$. Areas that are not included in this rotated box are defined as a negative area $A_n$. Moreover, we define areas neither positive nor negative as ignore, which is not considered during training. 

We define the pixels in the positive area $A_p$ as positive pixels and do the same to the negative and ignore area. Also, we set $k_{i, j} = 0$ for negative pixels and $k_{i, j} = -1$ for the ignore pixels. Due to the imbalance of the negative-positive pixels, we adopt focal loss as in \cite{lin2017focal}. The overall classification loss is the summation of focal loss over all non-ignoring pixels, which is normalized by the total number of positive pixels.
%

\myparagraph{Regression Branch}
The regression branch consists of a single 1x1 convolution, too. It outputs a regression map $\mathcal{R}$ of size $H\times W \times (4 + n \times 2)$. We will next elaborate on the meaning of each channel and the meaning of $n$. 




Ground truth 3D object from the bird's-eye-view is usually encoded as $(x, y, w, l, \theta)$. However, this encoding does not directly represent the relationship between each point inside the 3D objects with its corresponding boundary. As a result, we propose encoding ground truth 3D objects as $(\vec{b}, \theta)$, where the boundary vector $\vec{b}=(l,t,r,b)$ represents the distance from the positive pixel to the four sides of the corresponding bounding box (left, top, right, bottom), as illustrated in Figure \ref{fig:method-anchor-free} (A \textit{Car} example).


The total regression loss is computed as the average loss of positive pixels. For each positive pixel, the loss consists of two parts: IoU loss \cite{rezatofighi2019generalized} for $\vec{b}$, and bin-based rotation loss \cite{qi2018frustum} for $\theta$. We directly add up the weighted two parts to form the final regression loss. 

Specifically, for the loss of the boundary vector $\vec{b}$. We formulate the regression target for the boundary vector $\vec{b}_t=(l_t,t_t,r_t,b_t)$ in Equation \ref{eq2}.





\begin{equation}
\begin{aligned}
l_{t}= log(l), t_{t}=log(t)\\
r_{t}=log(r), b_{t}=log(b)
\end{aligned}
\label{eq2}
\end{equation}
we adopt IoU loss as in \cite{rezatofighi2019generalized} to optimize the boundary prediction in the regression output, which is formulated in Equation \ref{eq3}. $\vec{b}_p$ denotes the predicted bounding vector. 

\begin{equation}
\begin{aligned}
\mathcal{L}_{\vec{b}}=1-IoU(\vec{b}_p, \vec{b}_t)
\end{aligned}
\label{eq3}
\end{equation}



For the loss of orientation $\theta$, we observe that directly predicting orientation is hard. Thus we use a bin-based loss similar to that in \cite{qi2018frustum}, which decomposes the direct orientation regression into the bin classification and the residual regression within the corresponding bin. Specifically, we divide the orientation range of $2\pi$ into $n$ bins and define the bin target of the orientation in Equation \ref{eq4}. 

\begin{equation}
\begin{aligned}
\theta_t^{bin}=[\frac{\theta+\frac{\pi}{n}}{\theta_{n}}], \theta_{n}=\frac{2\pi}{n}
\end{aligned}
\label{eq4}
\end{equation}
Then the residual target in the corresponding bin is defined in Equation \ref{eq5}.

\begin{equation}
\begin{aligned}
\theta_t^{res}=\frac{2}{\theta_n}(\theta+\frac{\pi}{n}-(\theta^{bin}_t*\theta_n+\frac{\theta_n}{2}))
\end{aligned}
\label{eq5}
\end{equation}
The bin-based loss can thus be formulated in Equation \ref{eq6}. $\mathcal{F}_{\mathrm{cls}}$ represents the ctross-entropy classification loss, and $\mathcal{F}_{\mathrm{reg}}$ represents smooth L1 Loss \cite{girshick2015fast}.

\begin{equation}
\begin{aligned}
\mathcal{L}_{\theta} &= \mathcal{L}^{bin}_{\theta} + \mathcal{L}^{res}_{\theta}\\
&=\mathcal{F}_{\mathrm{cls}}\left(\operatorname{\theta}^{bin}_{p}, \operatorname{\theta}^{bin}_t\right) + \mathcal{F}_{\mathrm{reg}}\left(\operatorname{\theta}^{res}_{p}, \operatorname{\theta}^{res}_{t}\right)
\end{aligned}
\label{eq6}
\end{equation}
Finally, we define the overall regression loss in Equation \ref{eq7}.

\begin{equation}
\begin{aligned}
\mathcal{L}_{reg}&=\lambda_{\vec{b}} \mathcal{L}_{\vec{b}} + \lambda_{\theta} \mathcal{L}_{\theta}
\end{aligned}
\label{eq7}
\end{equation}
where $\lambda_{\vec{b}}$ and $\lambda_{\theta}$ is the weight for balancing two loss. We simply set $\lambda_{\vec{b}}=1$ and $\lambda_{\theta}=1$ here.


\begin{figure}
    \centering
    \begin{minipage}[t]{1.0\linewidth}
        \includegraphics[width=1\linewidth]{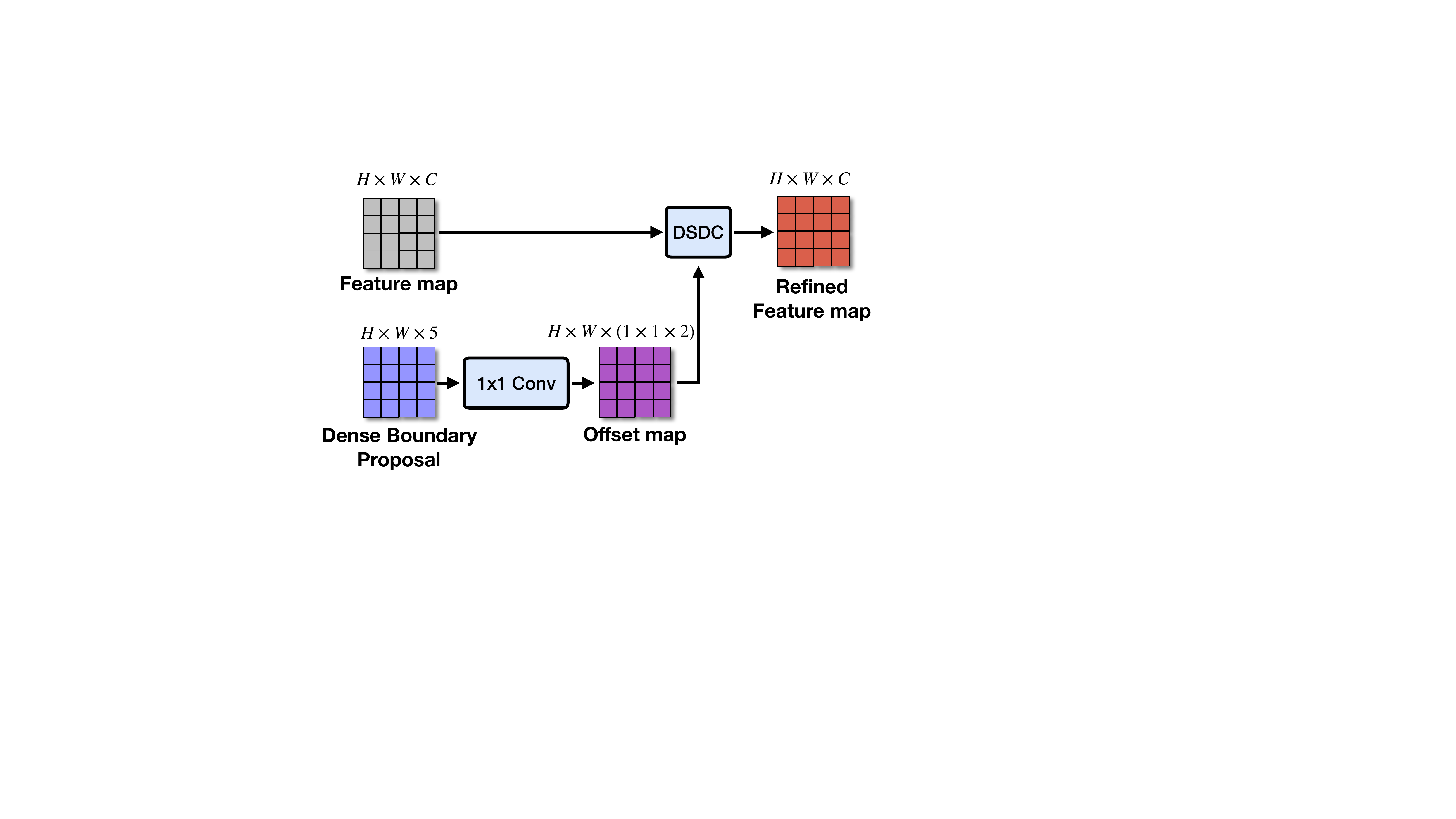}
    \end{minipage}%
    \caption{Illustration of DENFIConv. DSDC denotes depth-wise deformable convolution. The design of DENFIConv allows dense feature capture in a boudnary-aware manner.}
    \label{fig:denficonv}
\end{figure}

\myparagraph{Loss Function}
To sum up, we define our training loss function for DBPM in Equation \ref{eq10}:
\begin{equation}
\begin{aligned}
\mathcal{L}^{DBPM}&=\frac{1}{N_{pos}}\sum_{i, j}  \mathbbm{1}_{\left\{k_{i, j}\ge 0\right\}} \mathcal{L}_{cls}(p_{i, j}, k_{i, j}) \\&+ \lambda_{D} \frac{1}{N_{pos}}\sum_{i, j} \mathbbm{1}_{\left\{k_{i, j}>0\right\}} \mathcal{L}_{reg} (r^{*}_{p_{i,j}}, r^{*}_{t_{i, j}})
\end{aligned}
\label{eq10}
\end{equation}
where $\mathcal{L}_{cls}$ is the focal loss \cite{lin2017focal} and $\mathbbm{1}_{\left\{condition\right\}}$ is the indicator function (being 1 when $condition$ is true and $0$ otherwise). $r^*_p$ denotes the regression output $(\vec{b}_p, \theta_p^{bin}, \theta_p^{res})$, and $r^*_t$ denotes the regression target $(\vec{b}_t, \theta_t^{bin}, \theta_t^{res})$. $N_{pos}$ denotes the number of positive pixels and $\lambda_{D}$ is set to $1$ for balancing regression and classification loss. 

\myparagraph{Auto Scaling}
We also observe that it is difficult to directly regress the target in an anchor-free manner due to their extensive range, which can be monitored by generating anchor-free detection results from the classification branch and the regression branch of DBPM. Hence, we make use of a trainable scaling scalar $\dot s$ to automatically adjust the scale of the object information. As a result, instead of using the standard $r^*_p$, we make use of $\dot{s} \times r^*_p$ for the regression branch, which empirically improves the overall performance.

\myparagraph{Dense Boundary Proposal}
As shown in Figure \ref{fig:method-anchor-free}, we acquire the dense boundary proposal from the regression output of the DBPM. The classification branch of DBPM is only used in the training stage as an auxiliary task to ease the optimization of the regression branch.

Since the information from the regression output of DBPM is encoded to represents $(l_t, t_t, r_t, b_t, \theta_t^{bin}, \theta_t^{res})$, we decode it into original $(l, t, r, b, \theta)$ to form the dense boundary proposal. The dense boundary proposal is a tensor of size $H\times W\times 5$ serving as a pixel-wise indication for the boundary location. Note that we do not predict the information on the z-axis, including $(z, h)$, because we focus on deformable sampling on x-y dimensions. In this way, we make each pixel in the dense boundary proposal aware of the position of the boundary.





\subsection{Boundary-Aware Dense Feature Capture}
\label{subsec:fcaf-fc}
With the dense boundary proposal from DBPM, we aim at capturing dense feature in a boundary-aware manner. 


\myparagraph{Motivation}
Deformable convolution \cite{dai2017deformable} adjust the sampling position of the convolution layer dynamically by introducing an offset for each pixel of the feature map, thereby allowing us to guide the convolution layer where to focus. The offset map provides an offset of size $H\times W\times (F \times F \times 2)$ for each pixel of the feature map, where $F$ denotes the size of the convolution kernel, and $2$ denotes the offset of convolution sampling point along $H, W$ direction respectively. However, the original deformable convolution directly learns the offset map from the feature map, which is not accurate because the feature map does not explicitly provide the location of the dense features. We thus resort to explicit guidance to ease the optimization process. We also prove the significant advantage of explicit guidance in the ablation studies (See Table \ref{ablation:main}).


\begin{figure*}
    \centering
    \includegraphics[width=0.98\textwidth]{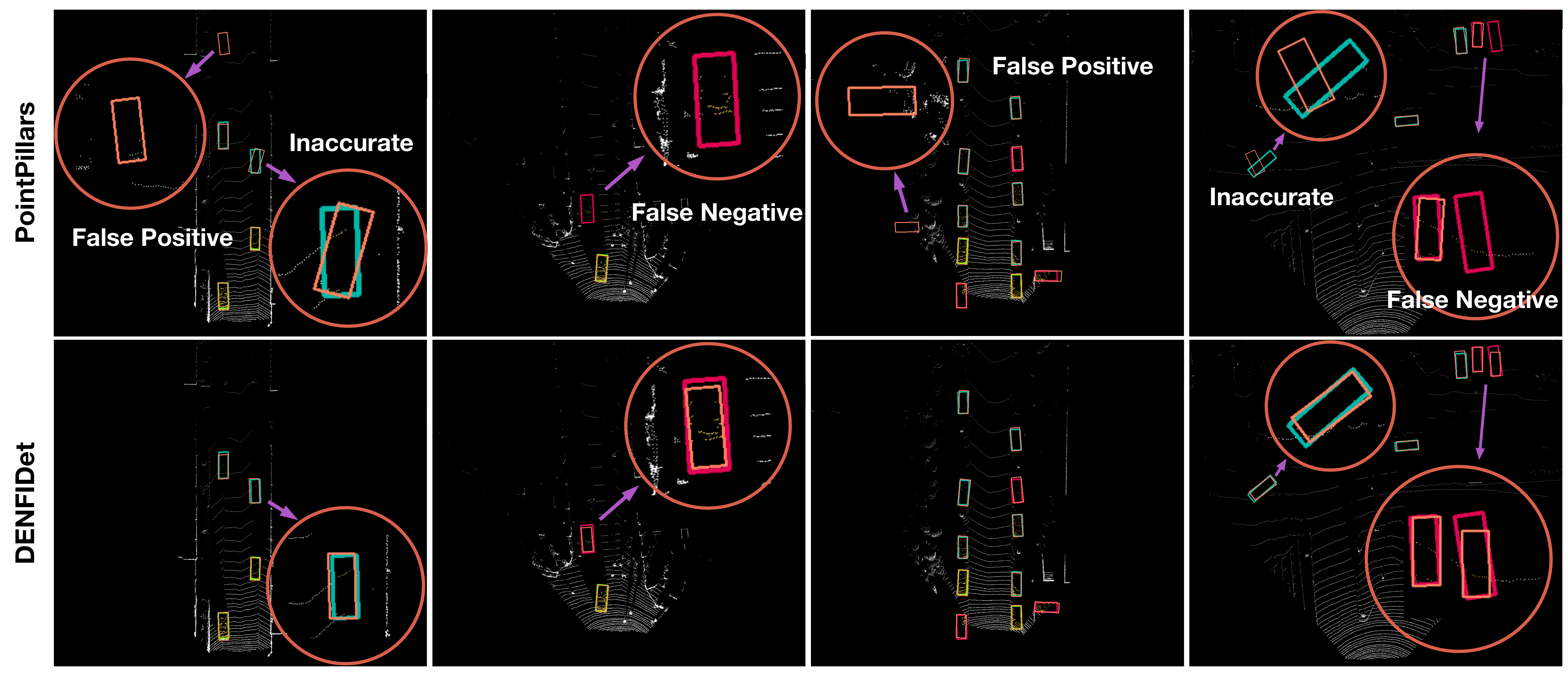}
    \centering    
    \caption{{Qualitative results of 3D object detection on the KITTI \emph{val} split}. \textbf{Top}: PointPillars, \textbf{Bottom}: Our DENFIDet. As a reference, we visualize the ground truth objects of easy, moderate, and hard levels as green, light blue, and red, respectively. We also visualize detection results as orange for comparison. The DENFI-guided 3D detector can detect objects more accurate and is good at capturing hard objects with fewer points. (Best viewed in color.)
}
    \label{fig:experiments-qualitative}
\end{figure*}

\myparagraph{DENFIConv}
As shown in Figure \ref{fig:denficonv}, we devise an operator called DENFIConv to utilize the boundary information for capturing dense features effectively. We learn the offset map by stacking 1x1 convolution on it since the dense boundary proposal contains pixel-wise boundary information. The offset map is then sent to a deformable convolution for explicit feature capturing on the backbone feature map. Finally, we use the refined feature map for detection purposes. We adopt a single DENFIConv for the regression branch and the classification branch of the detection head, respectively, which is shown in Figure \ref{fig:method}.

\myparagraph{Depth-Wise Separable Deformable Convolution}
\cite{wang2019region} adopts a 3x3 deformable convolution in their original paper for its sizeable receptive field. However, because the size of the backbone feature map is large, 3x3 deformable convolution brings significant computational overhead. Inspired by the design of MobileNets \cite{howard2017mobilenets}, we design depth-wise separable deformable convolution, which decomposes a 3x3 deformable convolution into a 3x3 depth-wise convolution and a 1x1 deformable convolution. This design delivers similar performance while running four times faster.

\subsection{DENFIDet}
\label{subsec:fcaf-fcaf}
We take PointPillars \cite{lang2019pointpillars} as our 3D object detection framework in DENFIDet. In short, PointPillars consists of three parts: a Pillar Feature Net for learning 2D compact tensor representation, a 2D convolutional Backbone for high-level feature learning, and an SSD detection head for 3D object detection and regression. We directly insert DENFI between the backbone and the SSD detection head to form DENFIDet.


\myparagraph{Joint Objective}
The loss function of DENFIDet consists of two parts. One part is from the anchor-based detection head of original PointPillars \cite{lang2019pointpillars}, which is denoted as $\mathcal{L}^{PP}$. Another part is from DBPM, which is $\mathcal{L}^{DBPM}$ defined in Equation \ref{eq10}. We train the network with multi-tasks loss defined in Equation \ref{eq100}. $\lambda$ is the weight parameter for balancing two tasks.

\begin{equation}
\begin{aligned}
\mathcal{L}=\mathcal{L}^{PP} + \lambda\mathcal{L}^{DBPM}
\end{aligned}
\label{eq100}
\end{equation}

%% file: sec_experiments.tex
\section{Experiments}
\label{sec:experiments}

\subsection{Dataset and Evaluation Protocol}
\myparagraph{Dataset}
We train and evaluate DENFIDet on the challenging KITTI \cite{geiger2012we} dataset. It contains 7481 training samples and 7518 testing samples with three categories (\textit{Car}, \textit{Pedestrian}, and \textit{Cyclist}). We use the same setting provided by \cite{zhou2018voxelnet,lang2019pointpillars} to divide the training data into \textit{train/val} split for experimental studies. The \textit{train} split contains 3712 samples and the \textit{val} split contains 3769 samples. For test submission, we create a \textit{mini-val} split containing 785 samples and train our model on the remaining 6696 samples as in \cite{lang2019pointpillars}.

\myparagraph{Evaluation Protocol}
Following \cite{lang2019pointpillars}, we evaluate the performance of 3D detectors on the KITTI bird-eye-view (BEV) benchmark and report the results of the new 40-point standard KITTI metrics (officially adopted by KITTI benchmark on October 8, 2019), which is the primary focus of KITTI 3D object detection task. Specifically, for each category of objects, they are divided into easy, moderate, and hard levels according to the heights of their 2D bounding boxes, occlusion levels, and truncation levels. We then calculate average precision (AP) with rotated IOU threshold 0.7 for \textit{Car} and 0.5 for \textit{Cyclist} \& \textit{Pedestrian} for each difficulty level. Besides, we report mean average precision (mAP) cross all categories.

\subsection{Implementation Details}
\label{subsec:experiments-details}

\myparagraph{Network Architecture}
We adopt the same settings as in PointPillars\cite{lang2019pointpillars} for point cloud transformation, anchors, and corresponding matching strategy for anchors. Please refer to the original paper for the details. Specifically, xy resolution is set to $0.16$ m. The focal loss in both DBPM and the detection head of PointPillars is set to $\alpha = 0.25$ and $\gamma=2$. For DBPM, we set $\sigma_1 = 0.3$, $\sigma_2=0.5$ for both \textit{Car} and \textit{Cyclist}, and $\sigma_1 = 0.5$, $\sigma_2=0.8$ for \textit{Pedestrian}. Moreover, we set the orientation bin num $n=12$; thus, we have orientation bin size $\theta_n=30^{\circ}$. Finally, we simply set $\lambda=0.5$ in the joint objective. Following \cite{lang2019pointpillars}, we train one network for \textit{Car} and one network for \textit{Pedestrian} and \textit{Cyclist}.


\begin{table*}[!tbh]
\centering
\scalebox{1.0}{
\begin{tabular}{c||ccc||ccc||ccc}
\multirow{2}{*}{Settings}             &  & $AP_{0.8}$&   & & $AP_{0.85}$& &  &$AP_{0.90}$ \\ \cline{2-10}
& $AP_E$ & $AP_M$ & $AP_H$  &  $AP_E$ & $AP_M$ & $AP_H$  & $AP_E$ & $AP_M$ & $AP_H$\\ \hline\hline 
      PointPillars \cite{lang2019pointpillars} & 79.56& 70.02& 66.80& 55.23& 46.48& 43.54 &15.75 & 13.41&12.15\\ \hline \\[-1em]

Ours (DENFIDet)     &  \textbf{82.42}&\textbf{72.56}&\textbf{68.41}&\textbf{60.86}&\textbf{50.90}&\textbf{46.58} &\textbf{18.58} & \textbf{16.05} & \textbf{14.01}\\ 

Relative improvement$\Uparrow$     &  \textbf{\color{JungleGreen}3.59\%}&  \textbf{\color{JungleGreen}3.63\%}&
        \textbf{\color{JungleGreen}2.41\%}& \textbf{\color{JungleGreen}10.19\%}&  \textbf{\color{JungleGreen}9.51\%}&  \textbf{\color{JungleGreen}6.98\%} &\textbf{\color{JungleGreen}17.97\%} &\textbf{\color{JungleGreen}19.69\%} &\textbf{\color{JungleGreen}15.31\%}\\ 

\end{tabular}}
\caption{Performance with increasing matching IoU threshold on KITTI \emph{val} split for \textit{Car} category. $AP_E$, $AP_M$, $AP_H$ denote the average precision on easy, moderate, hard levels respectively.}
\label{ablation:analysis}
\vspace{-3pt}
\end{table*}

\begin{table}[!tbh]
\centering
\scalebox{0.78}{
\begin{tabular}{c||ccc||c}
\multirow{2}{*}{Settings}             & \multicolumn{3}{c|}{mAP} & \multirow{2}{*}{Inference time}\\ \cline{2-4}
             &$AP_E$ &$AP_M$ & $AP_H$ &\\ \hline\hline 
      PointPillars \cite{lang2019pointpillars} &83.98& 73.68& 69.72          & 20ms  \\ \hline
PointPillars-DSDC      & 84.57&     73.95& 69.93&            \multirow{2}{*}{29ms} \\  

Relative Improvement$\Uparrow$      & \color{JungleGreen}0.70\%&  \color{JungleGreen}\color{JungleGreen}0.37\%&   \color{JungleGreen}0.30\%&           \\  \hline \hline
Ours (DENFIDet)     &  \textbf{86.81}  & \textbf{75.12} & \textbf{71.27}         & \multirow{2}{*}{29ms} \\ 
Relative Improvement$\Uparrow$      & \textbf{\color{JungleGreen}3.37\%}& \textbf{\color{JungleGreen}1.95\%}& \textbf{\color{JungleGreen}2.22\%}&       \\  \hline \hline

\end{tabular}}
\caption{Mean average precision on the KITTI \emph{val} split set for both \textit{Car}, \textit{Pedestrian} and \textit{Cyclists}. Performance of adopting different deformable modules is compared. DSDC denotes depth-wise separable deformable convolution. DENFIDet represents PointPillars-DENFI.}
\label{ablation:main}
\end{table}

\myparagraph{Training Details}
We implement the network with PyTorch \cite{paszke2017automatic} and adopt Adam optimization algorithm for training with a mini-batch size of 2. We train 160 epochs in total. The learning rate begins with 0.0002 and decays with a rate of 0.8 every 15 epochs. Considering the limited amount of training set on KITTI dataset, it is crucial to adopt data augmentation to alleviate the overfitting problem. We adopt the same data augmentation strategy as in PointPillars \cite{lang2019pointpillars} for a fair comparison. We first create a database containing all 3D ground truth bounding boxes and the point clouds falling inside theses 3D boxes. For each training sample, we randomly select 15,0,8 samples for \textit{Car}, \textit{Pedestrian}, and \textit{Cyclists}, respectively, from the database and place them into the current point cloud. Next, we randomly disturb ground truth objects by rotating (uniformly sampled from $-\frac{\pi}{20} \sim \frac{\pi}{20}$) and translating (x, y, and z independently sampled from $N(0, 0.25)$). Finally, for the global point cloud, we conduct random mirroring flip along the x-axis, global rotation (uniformly sampled from $-45^\circ\sim45$), global scaling (uniformly sampled from $0.95 \sim 1.05$) and global translation (x, y, z sampled from $N(0, 0.2)$).

\begin{table*}
    \centering
    {\small
\scalebox{0.81}{
\begin{tabular}{c|c|c||c|c|c||c|c|c||c|c|c||c|c|c}
\multirow{2}{*}{Method} & \multirow{2}{*}{Modality} & \multirow{2}{*}{Speed (ms)} & \multicolumn{3}{c||}{mAP}  & \multicolumn{3}{c||}{\textit{Car}} & \multicolumn{3}{c||}{\textit{Pedestrian}} & \multicolumn{3}{c}{\textit{Cyclist}} \\ \cline{4-15} \\ [-1em]
                        &                           &     &                      $AP_E$ & $AP_M$ &$AP_H$ & $AP_E$   & $AP_M$   & $AP_H$   & $AP_E$   & $AP_M$   & $AP_H$    & $AP_E$   & $AP_M$   & $AP_H$    \\ \hline \hline
               \textbf{Two-Stage Methods}            \\ 
         MV3D \cite{chen2017multi}           &      Lidar \& Img.             &      360                    & - & -&  -  &  86.62 &    78.93 &    69.80 &    -       &    -      &          -&     -     &      -   &  -       \\
         AVOD \cite{ku2018joint}       &     Lidar \& Img.              &       100        &  72.96& 65.09            &  59.23   &     89.75 &    84.95 &    78.32 &  42.58 &    33.57 &    30.14 &  64.11 &    48.15 &    42.37       \\ 
         AVOD-FPN \cite{ku2018joint}       &     Lidar \& Img.                 &       100       &  72.96& 64.09     & 59.23     &     90.99 &    84.82 &    79.62 &  58.49 &    \textbf{50.32} &    \textbf{46.98} &  69.39 &    57.12 &    51.09 \\ 
       F-PointNet \cite{qi2018frustum}      &         Lidar \& Img.                                   & 170  & 75.19 & 65.20 & 58.01   &    91.17 &    84.67 &    74.77 &   57.13 &    49.57 &    45.48 &    77.26 &    61.37 &    53.78      \\ 
        IPOD \cite{yang2018ipod}       &     Lidar \& Img.       &                            200  &  76.24 & 64.60 &  58.92   &     89.64 &    84.62 &    79.96 &     \textbf{60.88} &    49.79 &    45.43 &   78.19 &    59.40 &    51.38        \\ 

       F-ConvNet \cite{wang2019frustum}       &     Lidar \& Img.               &                            470  &77.57  & 67.89 & 60.16   &     91.51 &    85.84 &    76.11 &     57.04 &    48.96 &    44.33 &        \textbf{84.16} &    \textbf{68.88} &    60.05                \\ 
       UberATG-MMF \cite{liang2019multi}       &     Lidar \& Img                 &                               80  & - & -&  -    &     93.67 &    88.21 &    81.99 &       -       &    -      &          -&     -     &      -   &  -         \\ 
          STD \cite{yang2019std}    &        Lidar             &    80  &    \textbf{78.71} &    \textbf{68.38} & \textbf{63.44}   &    \textbf{94.74} &    \textbf{89.19} &    \textbf{86.42} & 60.02 &    48.72 &    44.55 &  81.36 &    67.23 &    59.35        \\ 
          Fast PointRCNN \cite{chen2019fast}    &        Lidar             &     65               & - & -         &   -    &  90.87 &    87.84 &    80.52 &
 -          &     -     &   -       &   -       &   -      &  -       \\ 
          PointRCNN \cite{shi2019pointrcnn}    &        Lidar            &     100               &76.49   & 66.92 &  61.95 &  92.13 &    87.39 &    82.72 & 54.77 &    46.13 &    42.84 &  82.56 &    67.24 &    \textbf{60.28}      \\ 
          \hline \hline
                \textbf{Single-Stage Methods}            \\
         ContFuse \cite{liang2018deep}       &      Lidar \& Img.            &           60   & - & -               &  -    &  \textbf{94.07} &    85.35 &    75.88 &
    -       &    -      &          -&     -     &      -   &  -      \\
                 HDNET \cite{yang2018hdnet}         &     Lidar \& Map        &       50      & - & -                &   -   &   93.13 &    87.98 &    81.23 &    -      &    -      &          -&     -     & -        &    -     \\ 

                PIXOR \cite{yang2018pixor}    &     Lidar            &                            35  & - & -&  -  & 83.97 &    80.01 &    74.31 &            -       &    -      &          -&     -     &      -   &  -        \\ 
               PIXOR++ \cite{yang2018hdnet}   &     Lidar        &                35      & - & -      &  -  & 93.28 &    86.01 &    80.11 &            -       &    -      &          -&     -     &      -   &  -         \\ 
             SECOND \cite{yan2018second}      &      Lidar            &            50      &  73.96 &  61.61      &      56.32 &     89.39 &    83.77 &    78.59 &     55.99 &    45.02 &    40.93 &          76.50 &    56.05 &    49.45       \\ 
             3D IOU Loss \cite{zhou2018iou}      &      Lidar            &            80      &  -  &   -      &  -     &     91.36 &    86.22 &    81.20 & -      &   -       &     -     &    - &    -     &   -      \\ 

                  PointPillars \cite{lang2019pointpillars}        &      Lidar   &         20       & 75.86 &  65.97       &  61.39    &    90.07 &    86.56 &    82.81 &           57.60 &    48.64 &    45.78 &   79.90 &    62.73 &    55.58        \\ 
            \textbf{Ours (DENFIDet)}   &          Lidar               &         29     & \textbf{78.57} &     \textbf{68.67}         & \textbf{63.52}    &  92.42 &    \textbf{88.56} &    \textbf{83.76} &
    \textbf{61.15} &    \textbf{51.96} &    \textbf{49.03} &
    \textbf{82.13} &    \textbf{65.49} &    \textbf{57.76}   \\ 
\end{tabular}
}
    }
    \caption{{Performance on KITTI \emph{test} BEV detection benchmark for both \textit{Car}, \textit{Pedestrian} and \textit{Cyclists}.} $AP_E$, $AP_M$, $AP_H$ denote the average precision on easy, moderate, hard levels respectively.
    }
    \label{table:experiments-kitti-test}
\vspace{-5pt}

\end{table*}

\myparagraph{Inference Details}
During inference, we first select the top 1000 detection results with the highest score from the output of the detection head. Then, we filter them with a score threshold of 0.05 and use rotated non-maximum suppression (NMS) with an overlap threshold of 0.01 IoU to generate final detection results. For fairness, both the speed of PointPillars and DENFIDet are measured in a PyTorch environment with a 2080Ti GPU and an Intel i7 CPU.

\subsection{Analysis Experiments}
\label{subsec:experiments-ablation}
In this section, we conduct detailed experimental studies to analyze the effectiveness of the introduced components. Results are evaluated on the KITTI \emph{val} split since the KITTI \emph{test} split can only be used for the submission of final results.

\myparagraph{Effect of Boundary-Aware Dense Features Capture}
\label{subsec:experiments-ablation-CFI}
In this part, we study the effectiveness of boundary-aware dense features capture for objects with a different scale. A straight baseline for our proposed module is the original depth-wise separable deformable convolution (denotes as DSDC). DSDC learns the offset map from the feature map itself while our method learns the offset map from the dense boundary proposal for explicit guidance. For a fair comparison, we make official PointPillar codebase\footnote{\url{https://github.com/nutonomy/second.pytorch}} as our another baseline. Then we add an extra deformable module on it, including both original depth-wise separable deformable convolution and DENFI. As shown in Table \ref{ablation:main}, we find that PointPillars-DSDC has only a minimal performance improvement compared to PointPillars. Compared to PointPilalrs-DSDC, DENFIDet outperforms it by \textbf{2.24, 1.17}, and \textbf{1.34 mAP} on the easy, moderate, and difficult levels with the same runtime. We also see more than \textbf{4}, \textbf{5}, and \textbf{7} times relative improvement on the easy, moderate, and hard difficulty levels compared to PointPilalrs-DSDC, which demonstrates the power of explicit boundary-aware feature capture, especially for those hard objects with fewer points on them.

%

\myparagraph{Runtime Analysis}
We next discuss the inference time of the different deformable modules. As shown in Table \ref{ablation:main}, they share the same computational overhead. The reasons are that \textbf{\texttt{1}}) DSDC uses the backbone feature map of size $H\times W\times C$ ($C=384$ in our experiments) to learn the offset map while DENFI uses the dense boundary proposal of size $H\times W\times 5$ for this purpose. \textbf{\texttt{2}}) we need to learn two offset maps for the classification and the regression branch of the anchor-based detection head. So even though DENFI has one more 1x1 convolutions for the regression branch of DBPM during inference, the runtime difference between DSDC and DENFI is negligible.

\myparagraph{Results Analysis}
In the previous experiments, we use an IoU threshold of 0.7 for the \textit{Car} category to calculate the AP for performance evaluation. However, detection with the highest possible quality is exceptionally critical in practical scenarios such as autonomous driving. In this section, we examine the effectiveness of DENFI on high-quality detection by increasing the IoU threshold of AP. The results are shown in Table \ref{ablation:analysis}. We observe that DENFI brings a much more significant performance improvement when the IoU threshold is set at a higher value for all difficulty levels. At the IoU threshold of 0.9, the relative improvement reaches \textbf{17.97\%, 19.69\%} and \textbf{15.31\%} on easy, moderate, and hard difficulty levels, which is $>$\textbf{5} times compared to that at the IoU threshold of 0.8. This demonstrates the effectiveness of DENFI for aggregating localization information for accurate detection. 



\subsection{Compared with State-of-the-Art}
\label{subsec:experiments-results}
We compare DENFIDet with a wide range of state-of-the-art 3D object detectors and summarize the results on the \textit{test} split. As shown in Table \ref{table:experiments-kitti-test}, compared to PointPillars\cite{lang2019pointpillars}, DENFIDet achieves better results across all categories and difficulty, outperforming it by \textbf{2.71}, \textbf{2.70}, and \textbf{2.13} \textbf{mAP} on easy, moderate and difficult levels respectively. In terms of performance improvement for different categories, DENFIDet surpass PointPillars by \textbf{2.00}, \textbf{3.32}, and \textbf{2.76} {\textbf{AP}} on the most critical "moderate" level for \textit{Car}, \textit{Pedestrian}, and \textit{Cyclist} categories respectively. It indicates that DENFI-guided 3D detection can achieve considerable performance improvement for objects with different scales and difficulty, especially for those hard objects with fewer points on them. 

With the significant improvement over the PointPillars\cite{lang2019pointpillars}, DENFIDet outperforms all methods, including both two-stage and multi-sensor fusion based, in terms of mean average precision (mAP), achieving new state-of-art performance with only point clouds as its input. Compared to previous best methods, DENFIDet is \textbf{2.7} times faster than STD \cite{yang2019std} (the prior art on KITTI) in speed and achieves better performance in a single-stage manner. Note that STD is a two-stage detector and needs to be trained stage-by-stage to save GPU memory while ours is trained end-to-end. Also, a significant performance advantage for the most difficult \textit{Pedestrain} category is observed, outperforming AVOD-FPN \cite{ku2018joint} (prior art for \textit{Pedestrain}. A multi-sensor and two-stage method) by \textbf{2.66}, \textbf{1.64} and \textbf{2.05 AP} on the easy, moderate and hard difficulty levels while running $>$ \textbf{3} times faster. This further demonstrates the power of the proposed DENFI module.

\subsection{Case Study}

Figure \ref{fig:experiments-qualitative} shows some qualitative comparison of baseline PointPillars and DENFIDet on the KITTI \emph{val} split set. As we see, PointPillars generate inappropriate results sometimes, including inaccurate regression, false-positive detection, and false-negative detection. Most of these faults are positioned in areas where points are very sparse, and thus it is easy to generate wrong results if we cannot capture their features accurately. In contrast, our DENFIDet performs well for those hard objects as DENFIDet always explicitly perceive where their densest features are.

%% file: sec_conclusion.tex
\section{Conclusion}

In this work, we pointed out the inconsistency of adopting 2D detection frameworks for 3D detection, which is caused by the uneven distribution of point clouds, and we can mitigate the problem by guiding 3D detector to explicitly capture the densest region in a boundary-aware manner. We propose a dense boundary proposal module (DBPM) to predict high-quality boundary information. We further design an efficient operator called DENFIConv to take advantage of the boundary information provided by DRBPM for dense features capture, thereby improving the quality of the features provided to the detection head. The simple and efficient design of DRBPM and DENFIConv allows us to combine them with many 3D detectors for improving performance while maintaining real-time speed.